\documentclass[conference]{IEEEtran}

\usepackage[T1]{fontenc}
\usepackage[utf8]{inputenc}
\usepackage{amsmath}
\usepackage{amssymb}
\usepackage{booktabs}
\usepackage{graphicx}
\usepackage{pgfplots}
\pgfplotsset{compat=1.17}
\usepackage[hidelinks]{hyperref}
\usepackage{url}

\begin{document}

\title{Replication Failure and Trivial Baselines in\\
Road-Level Crash Prediction}

\author{\IEEEauthorblockN{Maurya Patel}
\IEEEauthorblockA{\textit{School of Computer Science and Engineering} \\
\textit{University of Westminster}\\
London, United Kingdom \\
patelmaurya1112@gmail.com}}

\maketitle

\begin{abstract}
Graph neural networks are increasingly applied to road-level crash
prediction, but the stability of their reported gains has received little
scrutiny. We independently reconstruct the data pipeline of a recent
uncertainty-aware model and evaluate eleven of its design decisions across
three London boroughs under an expanding-window protocol. Four survive
replication on a second borough; seven do not, and four of those reverse
sign rather than attenuate. Multi-seed evaluation is decisive: one effect
reverses sign between random seeds within a single borough, and the
reference architecture exhibits per-borough seed spreads of up to 35.7
points against 4 points for ours. We further compare both networks against
a parameter-free baseline that ranks segments by cumulative past crash
count. At matched history depth our model is statistically
indistinguishable from that baseline ($-0.90$ points, $p=0.61$), and the
reference architecture loses to it on 18 of 18 held-out windows
($-17.37$, $p<10^{-6}$). Sweeping the baseline's lookback horizon shows it
spans 22.71\% to 83.94\% accuracy on that variable alone, and that every
published figure in this line of work is matched by the baseline at a
horizon of one to five years. We argue that the apparent margin of graph
networks over historical baselines in this task is substantially an
artefact of the short horizons those baselines were computed over, and
recommend horizon-matched baselines and multi-seed reporting as minimum
practice.
\end{abstract}


\section{Introduction}

Road traffic crashes remain a leading cause of death worldwide, and
accurate identification of high-risk road segments is a prerequisite for
targeted intervention. Recent work has applied graph neural networks
(GNNs) to this problem at road-segment granularity, reporting substantial
gains over statistical baselines~\cite{gao2024,nippani2023}.

Such gains are typically reported from a single training run on a small
number of regions. This is understandable given computational cost, but it
leaves two questions unanswered. First, do the design decisions that
produce the reported gains transfer to a second region? Second, are the
baselines against which those gains are measured constructed on comparable
terms?

This paper addresses both questions empirically. We reconstruct the data
pipeline of Gao et al.~\cite{gao2024} from public sources, evaluate it on
three London boroughs, and subject each design decision to replication on
an additional borough and to multi-seed evaluation. We then compare the
resulting model, and the reference architecture itself, against a
deliberately trivial baseline: ranking segments by their cumulative count
of past crashes.

The results are largely negative, and we argue that the negative results
are the more transferable contribution. Our specific contributions are:

\begin{itemize}
\item \textbf{A replication audit at scale.} Of eleven design decisions
that appear significant on one borough, four survive replication on a
second (Section~\ref{sec:replication}). Failure is characteristically a
sign reversal rather than an attenuation, which invalidates the common
assumption that a single-region estimate bounds the true effect.

\item \textbf{Evidence that effect size predicts non-replication only.}
Every effect below the measured seed-noise band failed to replicate, but
two effects well above it also failed. Effect size is therefore a
necessary but not sufficient filter.

\item \textbf{A horizon-matched trivial baseline.} At equal history depth,
our model is statistically indistinguishable from a parameter-free
crash-count sort, and the reference architecture loses to that sort on
every held-out window tested (Section~\ref{sec:baseline}).

\item \textbf{An explanation for the discrepancy with published results.}
Sweeping the baseline's lookback horizon shows a 61-point range on that
variable alone. Published historical baselines in this literature are
computed over single-year datasets, which we argue accounts for much of
the reported margin (Section~\ref{sec:horizon}).

\item \textbf{Measurement properties of the evaluation metric.} We
characterise the granularity of AccHR@20, show that training is not
deterministic at a fixed seed under standard GNN implementations, and
quantify the consequences for reproducibility (Section~\ref{sec:measure}).
\end{itemize}

\section{Related Work}

\subsection{Deep learning for road-level crash prediction}

Early crash-frequency models used linear and random-parameter regression
at area granularity. Recent work has moved to road-segment granularity
using spatiotemporal GNNs. Gao et al.~\cite{gao2024} propose STZITD-GNN, a
graph-attention encoder followed by a gated recurrent unit, decoded by a
zero-inflated Tweedie head to accommodate the extreme sparsity of
segment-level crash counts. Nippani et al.~\cite{nippani2023} assemble the
largest benchmark in this line, comprising nine million US crash records
with associated road networks and traffic volumes, and report that
GraphSAGE predicts monthly counts to within 22\% mean absolute error.

\subsection{The treatment of crash history}

A structural feature of this literature is that past crashes appear as the
prediction \emph{target} rather than as an input feature. In
Nippani et al.~\cite{nippani2023} the node features are graph-structural
(degree, betweenness centrality) together with weather and traffic volume,
and the reported leave-one-out ablation covers exactly those three
categories ($-6.9\%$, $-2.3\%$ and $-1.2\%$ respectively); crash records
enter as edge labels split temporally. Gao et al.~\cite{gao2024} include a
Historical Average baseline, but their dataset covers a single calendar
year, which bounds that baseline's lookback at one year.

By contrast, the Empirical Bayes method of the Highway Safety
Manual~\cite{hsm2010}, standard in transport practice, shrinks an observed
crash count toward a covariate-predicted mean and is conventionally
applied over three to five years of records, precisely because shorter
windows are too noisy at site level.

The gap this paper addresses is therefore not a modelling one. Both
communities recognise that past crashes predict future crashes; the deep
learning literature largely encodes this in the label, while the
safety-engineering literature encodes it in a multi-year feature.
Section~\ref{sec:horizon} quantifies what that choice is worth.

\section{Data and Method}

\subsection{Road network and target}

We use OS Open Roads (Ordnance Survey, Open Government Licence) as the
network source, clipped to each borough's administrative polygon by an
endpoint-inside test rather than a bounding box; the bounding box of
Westminster extends across the River Thames and admits approximately twice
the true link count. Segments are junction-to-junction links, and a
line-graph transform yields segment adjacency for message passing.
Westminster comprises 11,098 directed edges over 4,026 nodes.

Targets are daily collision counts per segment from STATS19 (UK Department
for Transport) for 2021--2024, snapped to the nearest edge (median
distance 2.3\,m; 0.22\% beyond 100\,m). At segment-day resolution the
target is 99.97\% zero.

\subsection{Features}

The model uses 35 features: segment geometry and node degree; day of week;
crash history over 7, 14, 30, 90 and 365 days and 2, 3 and 5 years;
casualty-type breakdown; annual average daily flow; four OpenStreetMap
point-of-interest density classes; and nine indices of multiple
deprivation. Long-horizon history is computed from the sparse collision
list through a cumulative-sum matrix rather than by rolling over the
segment-day table, which would require approximately $3\times10^{7}$
additional rows. This makes horizon length computationally free, which is
what permits the sweep in Section~\ref{sec:horizon}. Features are
standardised using statistics fitted on training instances only.

\subsection{Model}

A graph-attention layer is applied per timestep, and the resulting
embeddings are consumed by a gated recurrent unit (3 attention heads, 1
attention layer, hidden dimensions 16 and 32, residual connections). The
decoder is a zero-inflated Poisson head. Training uses Adam at learning
rate $0.01$ for 200 epochs without weight decay or early stopping.
Prediction intervals are obtained by split conformal calibration at the
90\% level on the final training instance.

\subsection{Evaluation protocol}

We use expanding-window walk-forward evaluation with a 20-day input
window, a 14-day prediction horizon and a stride of 90 days, giving six
held-out windows per borough spanning 2023-07-15 to 2024-10-07, with
training data growing from 6 to 11 instances. The feature table begins one
year before the instance grid so that the 365-day feature is complete for
every instance.

The evaluation metric is AccHR@20: for each day, the proportion of that
day's crashes falling on the top 20\% of segments by predicted risk,
averaged over the days of the window. This corresponds to the Acc@20
metric of the reference work.

\subsection{Statistical treatment}

All model figures are means over five random seeds. Comparisons are paired
window-by-window and report both a paired $t$-test and a Wilcoxon
signed-rank test. Where both arms of a comparison are model runs, pairing
is on seed as well as window, so that a seed's shared bias cancels.
Confidence intervals use the $t$ distribution ($n=5$, $t(4)=2.776$); the
normal approximation would understate them by approximately 40\%.

\section{Benchmark Comparison}

Table~\ref{tab:headline} reports our model against the figures published
by Gao et al. We emphasise that this is not a like-for-like comparison and
should not be read as one. The reference work evaluates within 2019 on a
6:2:2 split, whereas we use multi-year expanding walk-forward over
2022--2024. No paired test against the reference is possible, since only
three point estimates are published rather than per-window results. Our
target is also measurably sparser than reported (99.97\% zero against
95.72--96.71\%), a discrepancy we were unable to resolve after examining
the crash-rate formula, segment consolidation and evaluation protocol.

\begin{table}[t]
\caption{AccHR@20 over six walk-forward windows per borough, averaged over
five random seeds. Intervals use the $t$ distribution.}
\label{tab:headline}
\centering
\small
\begin{tabular}{lrrr}
\toprule
Borough & This work & 95\% CI & Gao et al. \\
\midrule
Westminster & $79.75 \pm 0.89$ & [78.64, 80.86] & 68.98 \\
Tower Hamlets & $82.75 \pm 2.13$ & [80.11, 85.40] & 72.24 \\
Lambeth & $77.75 \pm 1.67$ & [75.68, 79.82] & 76.59 \\
\midrule
Pooled & $80.08 \pm 2.62$ & [78.64, 81.53] & 72.60 \\
\bottomrule
\end{tabular}
\end{table}

\section{Replication of Design Decisions}
\label{sec:replication}

Each design decision that produced a significant effect on the first
borough was re-evaluated on a second. Table~\ref{tab:replication}
summarises the outcome.

\begin{table*}[t]
\caption{Replication outcomes for eleven design decisions. Four survived;
seven failed, of which four reversed sign.}
\label{tab:replication}
\centering
\small
\begin{tabular}{llll}
\toprule
Design decision & Effect (borough 1) & Second borough & Outcome \\
\midrule
Message-passing depth (1 vs.\ 2 layers) & $-26.46$ & $-43.21$ & Replicated \\
Encoder ordering (spatial vs.\ temporal first) & $-13.61$ & $-9.42$ & Replicated \\
Long-horizon crash history & $+15.85$ & $+9.70$ & Replicated \\
Proposed vs.\ reference architecture & $+18.53$ & $+13.51$ (pooled, 5 seeds) & Replicated \\
Hidden dimension $42/42$ & $+1.46$ & $-9.97$ & Failed \\
Architecture ensembling & $+1.79$ & $p=0.55$ at $n=18$ & Failed \\
Weight decay $0.01$ & $+2.53$ & $-0.60$ at $n=18$ & Failed \\
Road-class features harmful & $-6.87$ & $+2.25$ & Failed (sign reversal) \\
13 features equivalent to 35 & $-0.99$ & $-4.08$, then $+0.77$ & Failed (sign reversal) \\
Architecture $\times$ history interaction & $+25.04$ & Reverses sign & Failed (sign reversal) \\
Rank-transform feature scaling & $+5.80$ & $-39.70$ & Failed (sign reversal) \\
\bottomrule
\end{tabular}
\end{table*}

\subsection{Effect size predicts non-replication, not replication}

Effect size is frequently proposed as an informal filter for which
single-sample results to trust. On this evidence it operates in one
direction only. Every effect below the measured seed-noise band of
approximately 4 points failed to replicate (0 of 4), so a small effect is
reliable evidence \emph{against} replication. However, two effects
comfortably above that band also failed: road-class features ($-6.87$,
$p=0.016$) reversed sign on the second borough, and rank-transform scaling
moved from the best result recorded in our experiments ($+5.80$) to the
worst ($-39.70$). Large effect size is therefore necessary but not
sufficient evidence for replication.

\subsection{The form of failure matters}

Four of the seven failures were sign reversals rather than attenuations. A
practitioner assuming that effects merely shrink across regions --- and
therefore treating a single-region estimate as an upper bound --- would
have the direction wrong, not merely the magnitude. We regard this as the
more consequential finding, since it invalidates a common heuristic rather
than merely adding noise to it.

\subsection{Seed variance}

Multi-seed evaluation was decisive in two cases. Extending crash history
from five to nine years yields $-0.72$ points on one borough and $+2.98$
on another at a single seed. Across five seeds the effect reverses sign
\emph{between seeds within each borough} (Lambeth: $+0.27$, $+1.53$,
$-0.72$, $-1.79$, $-1.58$), and the pooled estimate is $-0.12$
($p=0.86$). No single-seed run of that experiment could have supported a
conclusion in either direction.

The reference architecture proved substantially more seed-sensitive than
ours, with per-borough spreads of 18.2, 20.8 and 35.7 points against
approximately 4 points for our configuration; one Westminster seed scored
42.58\%, barely twice the random baseline. Consequently its per-borough
disadvantage was mis-estimated in \emph{both} directions at a single seed
($+18.53$ against a true $+9.55$ on Lambeth; $+8.01$ against $+13.57$ on
Westminster). The direction of every comparison survived multi-seed
evaluation; none of the magnitudes did.

\section{Comparison Against a Trivial Baseline}
\label{sec:baseline}

We compare both networks against a parameter-free baseline that ranks
segments by cumulative past crash count, and against the Empirical Bayes
estimator of the Highway Safety Manual.

\begin{table}[t]
\caption{Mean AccHR@20 across three boroughs. Model figures are five-seed
means.}
\label{tab:baseline}
\centering
\small
\begin{tabular}{lr}
\toprule
Ranker & Mean AccHR@20 \\
\midrule
Crash-count sort ($\sim$8 years) & \textbf{83.94} \\
Empirical Bayes (HSM) & 83.84 \\
Crash-count sort, capped to 5 years & 80.99 \\
Proposed model (5 seeds) & 80.08 \\
Reference architecture (5 seeds) & 66.57 \\
\bottomrule
\end{tabular}
\end{table}

At matched history depth, our model and the parameter-free sort are
statistically indistinguishable ($-0.90$ points, $p=0.61$, 9 of 18
windows). Without the horizon cap, both trivial baselines exceed it
significantly (Empirical Bayes $-3.76$, $p=0.012$; crash count $-3.85$,
$p=0.023$).

This is not a property of our implementation alone. Evaluated on our data
at its own swept learning rate, with the same long history and averaged
over five seeds, the reference architecture loses to the crash-count sort
by 17.37 points on \emph{18 of 18} held-out windows ($p<10^{-6}$), and by
14.41 points against the horizon-matched sort. Neither graph network
tested in this study exceeds the trivial baseline.

\subsection{The baseline's strength is largely its horizon}
\label{sec:horizon}

Gao et al. report a Historical Average baseline at 44.96\% against their
model's 72.60\%, a margin of approximately 28 points that reads as clear
evidence for the network. Reconciling this with our sort's 83.94\%
requires only one variable: their dataset covers a single calendar year,
so their historical baseline can look back at most one year.

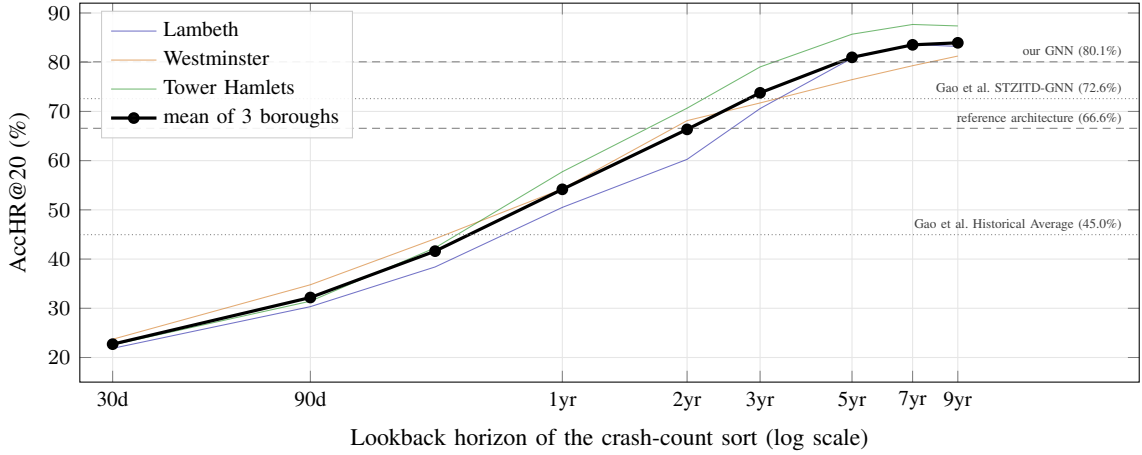
\begin{figure*}[t]
\centering
\begin{tikzpicture}
\begin{axis}[
  width=0.86\textwidth, height=6.6cm,
  xmode=log, log basis x=10,
  xlabel={Lookback horizon of the crash-count sort (log scale)},
  ylabel={AccHR@20 (\%)},
  xmin=25, xmax=9000, ymin=15, ymax=92,
  xtick={30,90,365,730,1095,1825,2555,3285},
  xticklabels={30d,90d,1yr,2yr,3yr,5yr,7yr,9yr},
  ytick={20,30,40,50,60,70,80,90},
  grid=major, grid style={gray!20},
  tick label style={font=\footnotesize},
  label style={font=\small},
  legend style={font=\footnotesize, at={(0.02,0.98)}, anchor=north west,
                draw=gray!40, fill=white, fill opacity=0.9, text opacity=1},
  legend cell align=left,
]
\addplot[blue!60!black, thin, opacity=0.55, mark=none] coordinates {(30,21.86) (90,30.33) (180,38.40) (365,50.50) (730,60.26) (1095,70.58) (1825,80.81) (2555,83.62) (3285,83.20)};
\addlegendentry{Lambeth}
\addplot[orange!80!black, thin, opacity=0.55, mark=none] coordinates {(30,23.71) (90,34.78) (180,44.13) (365,54.26) (730,68.13) (1095,71.72) (1825,76.46) (2555,79.29) (3285,81.25)};
\addlegendentry{Westminster}
\addplot[green!50!black, thin, opacity=0.55, mark=none] coordinates {(30,22.56) (90,31.44) (180,42.29) (365,57.74) (730,70.66) (1095,79.04) (1825,85.69) (2555,87.67) (3285,87.37)};
\addlegendentry{Tower Hamlets}
\addplot[black, very thick, mark=*, mark size=1.6pt] coordinates {(30,22.71) (90,32.19) (180,41.61) (365,54.17) (730,66.35) (1095,73.78) (1825,80.99) (2555,83.53) (3285,83.94)};
\addlegendentry{mean of 3 boroughs}
\addplot[gray, densely dotted, forget plot] coordinates {(25,44.96) (9000,44.96)};
\node[anchor=east, font=\tiny, gray!50!black] at (axis cs:8600,46.96) {Gao et al.\ Historical Average\ (45.0\%)};
\addplot[gray, densely dashed, forget plot] coordinates {(25,66.57) (9000,66.57)};
\node[anchor=east, font=\tiny, gray!50!black] at (axis cs:8600,68.57) {reference architecture\ (66.6\%)};
\addplot[gray, densely dotted, forget plot] coordinates {(25,72.60) (9000,72.60)};
\node[anchor=east, font=\tiny, gray!50!black] at (axis cs:8600,74.60) {Gao et al.\ STZITD-GNN\ (72.6\%)};
\addplot[gray, densely dashed, forget plot] coordinates {(25,80.08) (9000,80.08)};
\node[anchor=east, font=\tiny, gray!50!black] at (axis cs:8600,82.08) {our GNN\ (80.1\%)};
\end{axis}
\end{tikzpicture}
\caption{The same parameter-free crash-count sort, swept across lookback
horizons on our data and windows. The ranker spans 22.71\% to 83.94\% —
a 61.2-point range — with nothing changing but how far back it looks.
Dashed reference lines are measured on our data; dotted lines are the
reference paper's published figures on theirs, so their vertical position
is indicative rather than a controlled comparison.}
\label{fig:horizon}
\end{figure*}

Sweeping the same parameter-free ranker across lookback horizons on our
data (Figure~\ref{fig:horizon}) yields a range from 22.71\% to 83.94\%
with no change other than how far back it looks. Each published figure in
this line of work is matched by the sort at a short horizon: the
Historical Average at approximately one year, the reference architecture
at two, the reference model at three, and our own model at five.

This mapping is indicative rather than controlled, since two of those
figures were obtained on different data. What it establishes is narrower
but still substantial: the quantity of published improvement over a
historical baseline in this task is of the same order as the improvement
obtainable by extending that baseline's horizon by one to two years, using
data freely available in both cases. We identified no work in this line
that reports such a comparison.

The curve plateaus after approximately seven years, which corrects an
earlier reading of our own data: measured only to three years, it appears
to be climbing without plateau.

\subsection{Replication on an independent benchmark}
\label{sec:usrep}

The horizon result above rests on three boroughs of one city and
approximately $10^{4}$ crashes, which is the narrowest evidence base in
this paper. We therefore repeated the sweep on the ML4RoadSafety
benchmark~\cite{nippani2023}, using its Delaware subset: 458,282 crashes
over 166 months (2009--2022), falling on 36,466 distinct edges (those that
recorded at least one crash) of a full network of 109,107 undirected
edges, approximately 45 times the crash volume of the London data. That benchmark records monthly rather than daily counts, so the
sweep is over months and the metric averages over evaluation months; the
quantity measured is otherwise unchanged.

\begin{table}[t]
\caption{Horizon sweep on the Delaware subset of ML4RoadSafety at four
selection thresholds (AccHR, \%). Range is Best minus the 1-month value,
where Best is the maximum over all ten lookbacks, not necessarily the
10-year value. The dependence appears only where the task retains
headroom.}
\label{tab:usrep}
\centering
\small
\begin{tabular}{lrrrr}
\toprule
Threshold & 1 mo & 10 yr & Best (lookback) & Range \\
\midrule
Top 20\% & 87.75 & 93.04 & 93.04 (10 yr) & 5.3 \\
Top 5\% & 46.18 & 64.57 & 64.62 (7 yr) & \textbf{18.4} \\
Top 1\% & 16.16 & 30.10 & 30.63 (5 yr) & 14.5 \\
Top 0.2\% & 8.78 & 12.48 & 12.62 (3 yr) & 3.8 \\
\bottomrule
\end{tabular}
\end{table}

At the top-20\% threshold used throughout this paper the dependence is
only 5.3 points (Table~\ref{tab:usrep}), because a one-month lookback
already achieves 87.75\% and the task is close to saturated. At more
selective thresholds, where headroom remains, the dependence returns: 18.4
points at top-5\% and 14.5 at top-1\%.

The qualitative claim therefore replicates --- lookback horizon is a
first-order determinant of a crash-count ranker's accuracy, independent of
any model --- but its magnitude is regime-dependent, and the 61-point
figure obtained on London should not be read as general. We were unable to
establish the source of the magnitude difference. The most obvious
candidate, that the Delaware baseline starts high because a one-month
count leaves fewer segments tied at zero, is contradicted by measurement:
Delaware at one month leaves 97.6\% of edges tied against London's 94.2\%
at 30 days. Monthly versus fortnightly evaluation periods, crash
recurrence rates, network density and differing reporting thresholds
remain unseparated.

This does not affect the argument regarding single-year datasets. A
baseline computed on one year of data is bounded at a one-year horizon
whatever magnitude the horizon effect takes in that regime.

\section{Measurement Properties}
\label{sec:measure}

\subsection{Metric granularity}

AccHR@20 averages a per-day hit rate, so its smallest possible change
corresponds to a single crash crossing the threshold on a single day:
\begin{equation}
\Delta_{\min} = \frac{1}{c_d \cdot D}
\end{equation}
where $c_d$ is the number of crashes on day $d$ and $D$ the number of days
in the window. On our windows this ranges from 0.0102 to 0.0909. Since
over 94\% of segments are tied at zero recent crashes, a floating-point
difference invisible in the predictions can reorder a tie group and move
one crash across the threshold. No AccHR@20 figure on a sparse window
should be quoted more precisely than its own step size.

\subsection{Non-determinism at fixed seed}

Re-training a single window at a single seed three times produced 0.755,
0.832 and 0.845 --- a spread of 9.0 points --- while a control window on a
different borough was bit-identical across three repeats. Neighbourhood
aggregation in the attention layer is scatter-based, and CUDA does not fix
accumulation order; seeding makes initialisation reproducible but not
aggregation. We did not enable deterministic algorithms retroactively, as
this would break comparability with results already recorded, but
recommend it as the default for future work on this task.

\subsection{Divergence rather than degradation}

Three independent measurements found the same failure mode in the
reference architecture's design choices. Two-layer message passing
diverges on 2 of 6 Westminster windows; the full reference architecture
spreads 35.7 points across seeds; and the reference encoder ordering costs
approximately 6 points on four seeds but collapses to near-random accuracy
on the fifth, on the same seed in both boroughs tested. A mean over seeds
misdescribes all three cases, and the failure rate --- approximately one
in five here --- is the more informative quantity.

\subsection{Controls}

A constant prediction scores 10.12\%, below the random baseline of
approximately 20\%, confirming that no reported score is an artefact of
tie-breaking. Permuting targets across segments collapses the model from
75.64\% to 23.08\%, consistent with random. AccHR@20 additionally rises by
approximately 0.39 points per additional crash in a window ($p=0.0316$,
$n=31$), so sparse windows are measured less reliably; and holiday periods
are 11.14 points harder independently of crash volume ($p=0.0078$).

\section{Limitations}

The benchmark comparison covers three boroughs and one city, with six
held-out windows per borough. Generalisation was assessed on four
additional boroughs, across which the model ranges 73.83--84.66\% at a
single seed, indicating that performance does not collapse outside the
benchmark set. The target-density discrepancy against the reference work
remains unexplained.

Individual per-borough figures are less precise than their digits suggest.
We observed one window take three distinct values across three independent
runs of an identical configuration and seed, separated by exact
single-crash steps, while its five sibling windows were bit-identical. A
per-borough mean over six such windows may therefore shift by
approximately 0.3 points between runs for reasons unrelated to the model.

Data collection was affected by degradation of the OpenStreetMap Overpass
API, which reported free capacity while truncating larger transfers. One
borough's point-of-interest download lost an entire feature category on
four separate attempts while appearing successful; when eventually
retrieved intact, the truncated file was found to contain 24\% of the
actual data. Such responses are now rejected programmatically by
validating both category completeness and density against verified
downloads.

\section{Conclusion}

We reconstructed and evaluated a recent graph neural network for
road-level crash prediction, and find that most of its design decisions do
not replicate across regions. Of eleven decisions significant on one
borough, four survived replication; four of the seven failures reversed
sign rather than attenuating, which invalidates the assumption that
single-region estimates bound the true effect. Effect size predicts
non-replication reliably but replication only weakly.

More consequentially, neither the model we constructed nor the reference
architecture exceeds a parameter-free baseline that ranks segments by past
crash count. At matched history depth our model is statistically
indistinguishable from that baseline, and the reference architecture loses
to it on every held-out window tested. Sweeping the baseline's lookback
horizon shows that its strength is largely determined by that horizon, and
that published historical baselines in this literature are computed over
datasets too short to be competitive.

We therefore recommend that road-level crash prediction results be
reported against a horizon-matched crash-count baseline, and that per-seed
variability be reported rather than a single run. Both are inexpensive
relative to the cost of model development, and on this evidence either
would have changed the conclusions drawn.

\section*{Reproducibility}

All quantitative claims are regenerated by named scripts in the
accompanying repository, \url{https://github.com/Maurya1112-sudo/greyspot},
and two automated verifiers check the manuscript against its source data
and check the repository's documents for mutual consistency. Data sources
(STATS19, OS Open Roads, Indices of Multiple Deprivation, DfT annual
average daily flow) are public and free; none is redistributable, and
acquisition instructions are provided.


\begin{thebibliography}{9}

\bibitem{gao2024}
X.~Gao, X.~Jiang, J.~Haworth, D.~Zhuang, S.~Wang, H.~Chen, and S.~Law,
``Uncertainty-aware probabilistic graph neural networks for road-level
traffic crash prediction,'' \emph{Accident Analysis \& Prevention}, vol.
208, art.\ no.\ 107801, 2024, doi:10.1016/j.aap.2024.107801.

\bibitem{nippani2023}
A.~Nippani, D.~Li, H.~Ju, H.~N. Koutsopoulos, and H.~R. Zhang, ``Graph neural
networks for road safety modeling: Datasets and evaluations for accident
analysis,'' in \emph{Advances in Neural Information Processing Systems 36
(NeurIPS), Datasets and Benchmarks Track}, 2023.

\bibitem{hsm2010}
American Association of State Highway and Transportation Officials,
\emph{Highway Safety Manual}, 1st ed., Washington, DC, 2010.

\end{thebibliography}
\end{document}